\documentclass[review]{elsarticle}

\journal{Heliyon}

\usepackage{latexsym}
\usepackage{graphicx}
\usepackage{amsmath}
\usepackage{amssymb}
\usepackage{booktabs} 
\usepackage{array}
\usepackage{tabularx, multirow, booktabs}
\usepackage{subcaption}
\usepackage{algorithm}
\usepackage{algorithmic}
\usepackage{float}
\usepackage{comment}

\UseRawInputEncoding

\usepackage{enumitem}
\newlist{boldenum}{enumerate}{1}
\setlist[boldenum]{label=\textbf{(\arabic*)}}

\usepackage{xcolor}
\definecolor{newcolor}{rgb}{.8,.349,.1}

\begin{document}

\begin{frontmatter}

\title{On Preserving the Knowledge of Long Clinical Texts}

\author{Mohammad Junayed Hasan}
\author{Suhra Noor}
\author{Mohammad Ashrafuzzaman Khan\corref{corauth}}

\address{Department of ECE, North South University, Dhaka, Bangladesh.}

\cortext[corauth]{Corresponding author}
\ead{mohammad.khan02@northsouth.edu}






\begin{abstract}
Clinical texts, such as admission notes, discharge summaries, and progress notes, contain rich and valuable information that can be used for clinical decision-making. However, a severe bottleneck in using transformer encoders for processing clinical texts comes from the input length limit of these models: transformer-based encoders use fixed-length inputs. Therefore, these models discard part of the inputs while processing medical text. There is a risk of losing vital knowledge from clinical text if only part of it is processed. This paper proposes a novel method to preserve the knowledge of long clinical texts in the models using aggregated ensembles of transformer encoders. Previous studies used either ensemble or aggregation, but we studied the effects of fusing these methods. We trained several pre-trained BERT-like transformer encoders on two clinical outcome tasks: mortality prediction and length of stay prediction. Our method achieved better results than all baseline models for prediction tasks on long clinical notes. We conducted extensive experiments on the MIMIC-III clinical database's admission notes by combining multiple unstructured and high-dimensional datasets, demonstrating our method's effectiveness and superiority over existing approaches. This study shows that fusing ensemble and aggregation improves the model performance for clinical prediction tasks, particularly the mortality and the length of hospital stay. 
\end{abstract}


\end{frontmatter}

\section{Introduction}
Natural Language Processing is transforming the healthcare industry by providing practical ways to interpret and analyze the vast amount of clinical data \cite{dash2019big}. Usually, clinical data are stored as Electronic Health Records (EHRs) in massive databases. Texts in these records, specifically the admission and discharge notes, are rich in vital information that can revolutionize healthcare decision-making and improve clinical outcomes. Extracted knowledge from these notes can provide clues to doctors about courses of effective therapies and potential risks. Hospitals can use this data to predict capacity-related matters, such as patients' length of stay or in-hospital mortality. 

Harnessing the knowledge from the clinical data is a complex process. Efficient processing using NLP procedures requires a standard format of the data. Hospitals follow strict guidelines and formats while storing the data, but the rules and formats vary between hospitals and healthcare organizations. As a result, the clinical data is loosely structured, and the structure varies between organizations.  

Apart from the format, the data has information that is semantically distributed over time, irregular, and sporadic\cite{luo2016big}. The words and the vocabulary used in the records are highly domain-specific and carry human interpretation biases due to the complexity of the diseases and their conditions. This bias can sometimes meddle with the correct interpretations of the conditions or diseases. Getting help from the stored clinical data is only possible after dealing with these context-related interpretation issues. However, self-attention-based techniques (transformer-based language encoders) can effectively model the context in the languages and are well suited for clinical data. That is why transformer-based encoders/decoders whose power lies in finding representations with accurate context are being used to process medical data.  

Another area for improvement in clinical data processing is the length of the records. The clinical records keep on growing over the treatment period. These records hold the history of the medical care given to the patient and contain detailed descriptions of patient conditions, medications, procedures performed, and outcomes. The NLP model-building process must examine the entire record to find the correct context. Otherwise, the model will miss important information. However, the main limitation of the transformer-based encoders (for example, BERT) is their input mechanism: they can process only fixed-length inputs.  

The initial BERT model could process a maximum input sequence of length 512 tokens \cite{devlin2018bert}. Later, many transformer-encoder-based models used longer input sequences, but the length is always fixed. This limitation leads to context loss, performance degradation, and computational inefficiency \cite{beltagy2020longformer}. This problem limits the power and usefulness of advanced NLP models for solving real-world healthcare issues. If unresolved, crucial patient information could be lost or misinterpreted, adversely impacting clinical decisions. Solving this problem is essential for improving the quality and usability of clinical NLP systems and enabling them to handle more complex and diverse available data.  

This paper addresses this fixed-length input issue of the transformer encoders for medical text. We address long text inputs that exceed the capacity of most existing transformer-encoder models for improving the performance and robustness of clinical prediction tasks. 

Many studies have attempted to address this problem for transformer encoder models. Scientists usually use \textbf{text truncation} \cite{sun2019fine}, \textbf{text aggregation} \cite{pappagari2019hierarchical, su2021classifying, mahbub2022unstructured}, \textbf{text chunking} with a sliding window \cite{wang2019multi} or designing language models that are specific for long text inputs, such as Longformer \cite{beltagy2020longformer} and BigBird \cite{zaheer2020big}.  

\textbf{Text Truncation} discard the parts of the text that the model cannot handle, for example, anything more than 512 (or the max limit) tokens. The cut-short is done broadly in three ways: (i) Process the maximum length tokens from the beginning and discard the rest (ii) Process the maximum length tokens from the end and discard the tokens before (iii) Systematically select the most important tokens from the text and discarding the rest. \textbf{Text Aggregation} splits texts into multiple segments (each with a length equal to the maximum allowed input length) and then classifies each segment separately. Then aggregates the results using different strategies such as hard voting, majority voting, soft voting, averaging, or weighted averaging. \textbf{Text Chunking} with sliding window splits long texts into multiple segments by keeping overlaps between segments to drag the information from one segment to another. Then, each segment is classified separately or aggregates their outputs using various strategies. Language Models like Longformer and BigBird extend the maximum input sequence length from 512 to 4096 tokens. However, the problem persists as many records may exceed this fixed length.  

Although these three techniques have various advantages and benefits, they only partially solve the problem. We identify some crucial shortcomings and limitations here. \textbf{(a) Critical Information Loss:} Methods like truncation and using pre-trained language models like Longformer and BigBird may discard important information in the text's truncated parts. This leads to information loss since the text is truncated beyond 512 or 4096 tokens. Clinical texts and admission notes usually contain more than the mentioned number of tokens, hence being prone to critical information loss during training and inference. \textbf{(b) Contextual Discontinuity:} Techniques such as text chunking with a sliding window, while aiming to preserve continuity, can introduce breaks in the contextual flow of information. If significant information strides the boundary between two segments, the model might fail to recognize the relationship between the two pieces of information. \textbf{(c) Annotation Challenges:} Handling long clinical texts requires high-quality annotations for training supervised models. The sheer length and complexity of these texts can make the annotation process cumbersome, time-consuming, and prone to errors, affecting the overall quality of the trained model. \textbf{(d) Domain-specific Challenges:} Clinical language is laden with domain-specific terminologies and abbreviations and often presents itself in a non-standard form. Techniques that work well on general texts might not necessarily perform effectively on clinical notes, exacerbating the problem of long-text handling in clinical NLP. 
\begin{figure}[h]
	\centering
	\includegraphics[scale=0.58]{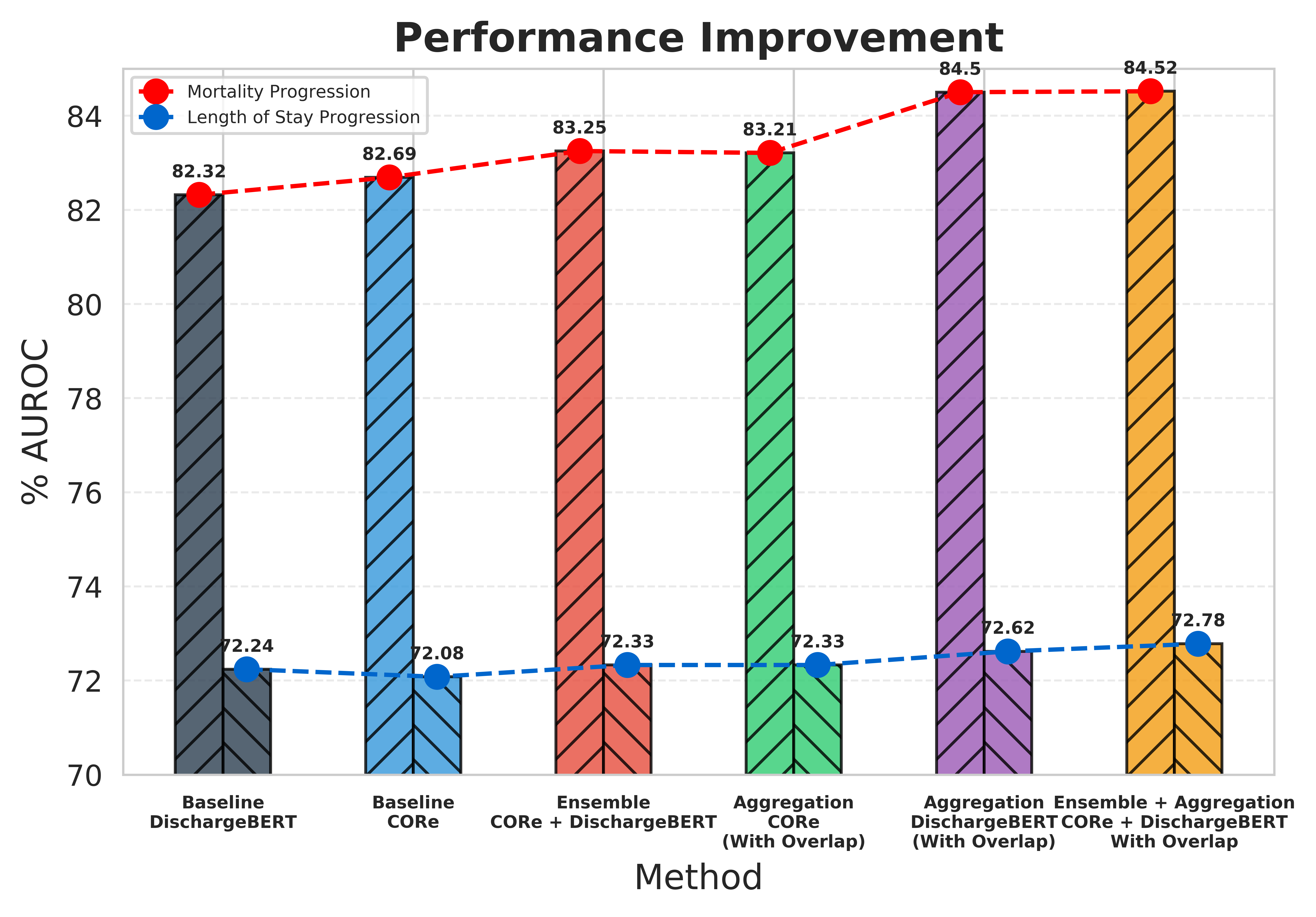}
	\caption{Average Performance Improvement from baseline approach to ensemble + aggregation approach as followed in our methodology in macro averaged \% AUROC for the clinical tasks Mortality Prediction and Length of Stay Prediction using the admission notes from MIMIC-III clinical database with context constrained large language models like BERT and its derivatives.}
	\label{fig:progress}
\end{figure}
\textbf{(e) Scalability Issues:} As the size and complexity of clinical databases continue to grow, the methods to handle long texts need to be scalable. While effective on smaller datasets, many current solutions might not scale well to larger, more diverse datasets without a decline in performance or a significant rise in computational costs. In light of these challenges, there is a pressing need for more comprehensive and scalable solutions that can effectively handle long clinical texts while ensuring accuracy, efficiency, and the preservation of critical information.

This paper explores two systematic ways of processing the whole text blocks. Our techniques address the issues we just discussed in the previous paragraph. We use sliding window text aggregation and model ensemble. Our approach uniquely combines ensembling and aggregation in contrast to leveraging them individually. Figure \ref{fig:progress} shows the average improvement in the predictive performance of the methodology followed in this study compared to existing methods like ensembling and aggregating in two crucial clinical tasks: mortality prediction and length of stay prediction. This synergistic combination ensures we harness each technique's benefits while offsetting limitations. For instance, where ensembling boosts the model's robustness, aggregation assists in handling the extensive lengths of clinical texts. To further the robustness of our approach, we collect and perform extensive experiments on the admission notes from the MIMIC-III clinical database and show that our method outperforms strong baseline approaches. 

\textbf{Contributions.} The main contributions of this study can be summarized as follows:

\begin{boldenum}[itemsep=-3pt]
	\item We explored the possibility of using an ensemble of transformer-based encoders to process long text. Processing long sequences of text is essential for clinical text as information critical for patient care could remain in any place of the medical records. Most other works in this area discard most of the text.

	\item We fine-tune multiple pre-trained large language models on two important clinical tasks, mortality prediction and length of stay prediction, and show that our method outperforms the baselines and current state-of-the-art.
	
	\item The proposed approach is designed to be scalable, accommodating the ever-growing sizes and complexities of clinical databases without compromising performance. Our integrated approach shows that we can substantially minimize the loss of critical information in clinical texts.
	
	\item We conduct extensive experiments on the admission notes from the MIMIC-III clinical database and provide a comprehensive analysis of our results.
	
	\item We ensure transparency and foster further research by publicly making our code, pre-trained models, and implementation details accessible.
\end{boldenum}

\section {Dataset}

We used the Medical Information Mart for Intensive Care III (MIMIC-III) clinical database for this work. It is a publicly accessible and comprehensive database with de-identified health-related data of forty thousand patients admitted to critical care units at the Beth Israel Deaconess Medical Center between 2001 and 2012 \cite{johnson2016mimic}. This dataset has information about the patient's vital sign readings, tests, imaging, lab reports, procedures performed, medications, hospital notes, outcomes, demographics, etc.   

Clinical notes recorded by healthcare professionals provide intricate narratives on patient conditions, interventions, and progress. These notes connect the information present in the medical records. Predictive modeling using such crucial information can potentially extract nuanced patterns of diseases, symptoms, etc. Furthermore, the massive volume of information and diverse patient population within MIMIC-III ensures a broad representation, deriving more generalizable and robust model inferences.

\subsection{Clinical/Admission Notes}

Clinical notes (here the admission notes) of this dataset is pivotal to the healthcare domain. They contain detailed information of patient's condition, history, and the healthcare services rendered. However, we used the following information to build our models.

\begin{itemize}
	\item Chief Complaint ($CC$): The primary reason or concern for which the patient seeks medical attention.
	\item Present Illness ($PI$): A detailed account of the symptoms and problems leading up to the current visit, typically in chronological order.
	\item Medical History ($MH$): A comprehensive record of the patient's past diseases, surgeries, and medical interventions.
	\item Admission Medications ($AM$): The list of medications the patient was taking at the time of hospital admission.
	\item Allergies ($AL$): Documentation of substances that cause adverse reactions in the patient, such as drugs, food, or environmental elements.
	\item Physical Exam ($PE$): Observations from the physician's examination of the patient, including vitals, appearance, and specific tests.
	\item Family History ($FH$): Information about health conditions and diseases that run in the patient's family, highlighting potential genetic risks.
	\item Social History ($SH$): Details about the patient's lifestyle, habits, occupation, and social activities, which might influence health outcomes.
\end{itemize}

The above information is divided into columns in the dataset. We extracted certain columns from these notes and represented them as chunks of text for further processing. Given a set of clinical/admission notes, each note $N_i$ is represented as:

\begin{equation}
	N_i = CC_i \oplus PI_i \oplus MH_i \oplus AM_i \oplus AL_i \oplus PE_i \oplus FH_i \oplus SH_i 
\end{equation}

where $\oplus$ represents the concatenation operation. After this basic representation, the notes undergo a two-step preprocessing phase.

\subsection{Data Preprocessing}

We extracted the clinical/admission notes from the MIMIC-III clinical database to predict mortality and length of stay. This database is popular in the research community. However, we were careful to use a different subset of data from this database when we also used pre-trained models (trained using the same database but a different part of the dataset). For example, \cite{van2021clinical} used admission and discharge notes from this database, while we used different columns from the admission notes. Then, we used the data from these columns to make stories about the patients' treatments. 

Specifically, we used data from two files of the clinical admission notes: \texttt{noteevent.csv} and \texttt{admission.csv}. We performed strategic concatenation of specific columns from these files. The columns of interest are Chief Complaint, History of Present Illness, Medical History, Admission Medications, Allergies, Physical Exam, Family History, and Social History.  

These columns are merged to build a unified column - termed \textit{text}. This newly formed \textit{text} effectively encapsulates comprehensive clinical notes, offering an integrated narrative of the patient's medical trajectory of the treatment process. 

After that, we split the data into train and validation sets. This data processing phase was pivotal in translating raw clinical notes from MIMIC-III into a dataset with medical processes described over time. This derived data provides a comprehensive portrait of patient admissions, laying the groundwork for our subsequent predictive modeling tasks.

\begin{table}
	\footnotesize
	\caption{The train-validation-test split of the extracted clinical admission notes from the MIMIC-III clinical database on two clinical tasks: mortality prediction and length of stay prediction. The split is done in a 70:10:20 ratio.}
	\begin{tabular}{lcccc}
		\toprule
		\tiny
		& \multicolumn{4}{c}{\textbf{Number of Samples}} \\
		\cmidrule{2-5}
		\textbf{Task} & \textbf{Total Extracted} & \textbf{Train Set} & \textbf{Validation Set} & \textbf{Test Set} \\
		\midrule
		Mortality Prediction & 48684 & 33954 & 4908 & 9822 \\
		Length of Stay Prediction & 43609 & 30421 & 4391 & 8797 \\
		\bottomrule
	\end{tabular}
	\label{tab:datasets}
\end{table}

These are the major steps from preprocessing to building the model of this research:

\begin{enumerate}
	\item \textbf{Processing Clinical/Admission Notes} $(N$): This is the first step to preparing the input data for the model. We represent the interesting \textit{text} from the notes as a sequence of tokens, $N = \{n_1, n_2, ..., n_m\}$, where $m$ is the total number of tokens. $N$ encapsulates comprehensive patient data encompassing medical histories $H$, symptom vectors $S$, diagnostic outcomes $D$, and treatment protocols $T$.
	
	\item \textbf{Tokenization}: The input \textit{text} (from the step above) $N$ undergo a tokenization process, resulting in a set $T_N = \{t_1, t_2, ..., t_m\}$ where $m$ is the length of the note in tokens. 
	
	\item \textbf{Text Chunking}: The tokens are then segmented into chunks, $C = \{c_1, c_2, ..., c_k\}$ such that each chunk $c_i$ has a maximum length of 512 tokens with an overlap of 50 tokens. To maintain semantic continuity with BERT-like models, each chunk $c_i$ is appended with [CLS] and [SEP] tokens, denoted as $c_i^{'}$.
	
	\item \textbf{Baseline Model Ensemble}: Let $B = \{b_1, b_2, ..., b_p\}$ represent the ensemble of $p$ BERT-based baseline models (pre-trained on related medical text). Each chunk $c_i^{'}$ is passed through the model $b_j$ in $B$. The prediction generated by model $b_j$ for chunk $c_i^{'}$ is denoted as $P_{i,j}$. Thus, the prediction matrix for all chunks and models is represented as $P = \{P_{1,1}, P_{1,2}, ..., P_{k,p}\}$.
	
	\item \textbf{Prediction Aggregation}: The predictions from each chunk $c_i^{'}$ are aggregated using a weighted average mechanism. Let $W = \{w_1, w_2, ..., w_k\}$ be the set of weights for each chunk. The aggregated prediction $A_i$ for chunk $c_i^{'}$ is given by:
	\begin{equation}
		A_i = \sum_{j=1}^{p} w_j \times P_{i,j}
	\end{equation}
	
	The final prediction $A$ for the clinical note $N$ is then obtained by consolidating all $A_i$, typically through averaging or another fusion strategy.
\end{enumerate}

The inclusion of [CLS] and [SEP] tokens is crucial because BERT-based models, by design, expect these tokens. The [CLS] token is an aggregate representation of the entire input, suitable for classification tasks. Especially in clinical notes, the beginning and end of a chunk can be mid-sentence. The [CLS] and [SEP] tokens offer boundary demarcations, ensuring the model does not misinterpret the chunk's context. During pre-training, BERT models are trained with these tokens. Omitting them during fine-tuning or inference may lead to suboptimal performance. The tokenization and chunking approach, augmented with [CLS] and [SEP] tokens, ensure that each segment of the clinical note is contextually intact, facilitating the extraction of coherent embeddings and, thus, robust predictions. This rigorous preprocessing regimen ensures the structured integration of diverse clinical data into a unified format, ready for downstream analysis by aggregation and the ensemble of models.

\section{Related Works}
\textbf{Context Constrained LLMs.} BERT (Bidirectional Encoder Representations from Transformers) used transformer architecture \cite{vaswani2017attention} and implemented bidirectional context extraction from text. The success of the transformers\cite{vaswani2017attention} and BERT in solving several core NLP tasks efficiently\cite{devlin2018bert} triggered a series of transformer-based successors. However, most of these works kept the input size limited to 512 tokens. RoBERTa \cite{liu2019roberta}, an enhancement over BERT, fine-tuned the training methodology and data to achieve superior performance but did not address the 512-token limitation. DistilBERT \cite{sanh2019distilbert} reduced the size of BERT to improve performance and also did not address the input token limitation. ALBERT \cite{lan2019albert} innovated structurally by factorizing the embedding layer and sharing parameters across layers, reducing the model's size without compromising its efficacy. Still, the 512-token context limitation remained. 

One of the early research that addressed the input-token constraint was Longformer \cite{beltagy2020longformer}. It used local and global attention mechanisms, pushing the envelope by efficiently managing documents with 4096 tokens. Similarly, BigBird \cite{zaheer2020big} incorporated sparse attention patterns to process extended sequences. However, their design entails predefined global tokens, potentially limiting their applicability to specific tasks or domains. Moreover, despite their advancements, a maximum sequence length constraint persists, which becomes evident when processing extended clinical texts. 

Clinical texts entail unique linguistic complexities; thus, domain-specific models emerged for related processing. ClinicalBERT \cite{huang2019clinicalbert} is a BERT adaption, fine-tuned on the MIMIC-III database's clinical notes, ensuring fine clinical language representation. BioBERT \cite{lee2020biobert}, although not strictly clinical, was honed for biomedical tasks. Similarly, BlueBERT \cite{peng2019transfer}, focusing on the biomedical domain, used embeddings tailored to PubMed abstracts. MedBERT \cite{rasmy2021med}, targeted at medical transcriptions, captures the essence of patient-doctor interactions, while BioDischargeSummaryBERT \cite{alsentzer2019publicly}, pre-trained on discharge summaries, emphasizes domain-specific embeddings' superiority over generic ones. CORe \cite{van2021clinical}, evolving from BioBERT, emphasizes patient outcomes, setting new benchmarks for clinical outcome predictions. 

Recent research like \cite{li2022clinical} fine-tuned models such as Longformer and BigBird on clinical texts, yielding Clinical-longformer and Clinical-bigbird, capable of handling 4096 tokens. Despite these endeavors, the fundamental context size constraint of 512 or 4096 tokens—remains an unaddressed challenge in BERT-like language models.

{\textbf{Long Text Handling.}} Several studies explored different methods for processing long input using transformer encoders (like BERT). Central to this endeavor are approaches like text aggregation and overlapping sliding windows, both tailored to ensure contextual continuity across divided chunks of lengthy documents. To start, \cite{pappagari2019hierarchical} harnessed a hierarchical methodology to address this challenge. Their strategy segmented the input into more digestible portions, each being independently processed through a base model. The outputs of these chunks underwent aggregation via either a recurrent layer or an additional transformer and then to softmax activation. This sequence ensures that the intrinsic meaning within the divided segments is not lost during processing. Similarly, \cite{su2021classifying} adopted a text aggregation mechanism. Opting for the DistilBERT model, their approach was particularly designed for expansive clinical texts, a nod towards the computational and resource constraints of earlier models. Offering another perspective, \cite{mahbub2022unstructured} relied on an averaging technique to generate embeddings from various clinical text fragments. These were then utilized in more traditional machine learning architectures, such as Random Forest and feed-forward neural networks. Their end goal was to predict the length of a patient's stay based on distinct time frames. Highlighting the merits of overlapping windows, \cite{wang2019multi} demonstrated the efficacy of their chunking technique. They could substantially outperform conventional methods in their experiments by dividing texts into smaller passages and processing them with a sliding window. Similarly, \cite{su2021classifying} innovatively integrated a sliding selector network with dynamic memory to succinctly summarize extensive documents. \cite{grail2021globalizing} offered a layered perspective through their hierarchical sliding window strategy. Their approach allowed inputs to be processed blockwise using scaled dot-attentions, with subsequent layers intelligently integrating these processed blocks. 

Parallel to these focused methods, some researchers pursued the refinement of transformer architectures, as seen in works like \cite{dai2019transformer, kitaev2020reformer, rae2019compressive, sukhbaatar2019adaptive}. Among them, \cite{qiu2019blockwise} stands out by tailoring BERT to incorporate sparse block structures within the attention matrix. This modification streamlined memory usage and computation times while empowering attention heads to discern short- and long-term contextual nuances. However, the journey to refine long text handling has its pitfalls. Recognizing inherent drawbacks like insufficient long-range attention or the specialized computational requirements in some techniques, \cite{ding2020cogltx} steered their research towards a dual-model system. By jointly training two BERT (or RoBERTa) models, they crafted a mechanism to cherry-pick pivotal sentences from lengthy documents. This strategy has proven invaluable across various tasks, including text classification. In summation, while significant strides have been made in handling extensive texts with LLMs, an intricate dance exists between segmenting these texts and preserving their inherent context. Our research seeks to perfect this intricate balance, harmonizing text segmentation, context preservation, and optimized performance.

{\textbf{Ensembles of BERT in Clinical NLP.}} The transformative influence of the BERT architecture, derived from Transformer models, has significantly enriched the landscape of Clinical NLP. A notable shift from traditional tokenization methods was observed in \cite{boukkouri2020characterbert}, where CharacterBERT opted for a character-centric approach instead of the standard wordpiece tokenization. This was a move towards conceptual simplicity and showed how ensemble models could be deployed using a majority voting strategy for enhanced outcomes. Another significant stride was made by \cite{amin2019mlt}, which unveiled the power of transfer learning in Clinical NLP. Their approach utilized BERT and its specialized variant, BioBERT, in an ensemble framework, establishing new benchmarks. BERT's ensemble strength was further spotlighted in diverse applications: \cite{dang2020ensemble} harnessed it to effectively detect medication mentions from Twitter data. In contrast, \cite{senn2022ensembles} demonstrated its effectiveness in detecting depression signs from clinical interview transcripts. Going beyond the standard application of BERT, \cite{lin2020does} integrated it with neural unsupervised domain adaptation techniques, opening avenues for enhanced negation detection in clinical narratives. 

The challenge of language specificity in NLP tasks was tackled ingeniously by studies \cite{li2020chinese} and \cite{zhou2023ensemble}. Both emphasized the potency of ensemble strategies in addressing named entity recognition challenges in Chinese clinical records, setting new performance standards. Lastly, medication information extraction, vital for healthcare, significantly advanced with \cite{kim2020ensemble}. Their work, built on a stacked ensemble, not only excelled in extracting adverse drug events and medication details but also confirmed the broad-spectrum potential of BERT-based ensembles across varied clinical NLP challenges. Building on the ongoing evolution and applications of ensemble methods in this domain, we first leverage ensemble modeling for two clinical tasks: mortality prediction and length of stay prediction.

{\textbf{Prediction of In-Hospital Mortality and Length of Stay.}} Many recent research studied the effectiveness of machine learning approaches for predicting mortality and the length of stay from clinical texts. Broadly, these techniques can be categorized based on their architectural and theoretical foundations. Hierarchical designs are evident in several studies aiming to extract a layered understanding from clinical texts. \cite{feng2020explainable} employed a hierarchical CNN-transformer model for ICU patient mortality prediction in a multi-task setting. \cite{si2020patient} took a similar hierarchical approach using attention-based RNNs, focusing on distinguishing meaningful gaps between notes. \cite{zhang2020time} proposed a time-aware transformer-based hierarchical structure specifically for mortality prediction. The power of transformer architectures, especially BERT, has been harnessed by many for these tasks. \cite{zhao2021bertsurv} introduced a deep learning survival framework leveraging BERT for language representation. \cite{darabi2020taper} utilized transformer networks like BERT to embed clinical notes to predict mortality and length of stay. Moreover, \cite{van2021clinical} combined self-supervised knowledge integration and clinical outcome pre-training using admission notes for similar predictions. Incorporating diverse data types, \cite{deznabi2021predicting} combined clinical notes with MIMIC- II's time-series data for in-hospital mortality prediction. \cite{yang2021multimodal} championed a temporal-clinical note network for this purpose, while \cite{jana2022using} concentrated on nursing notes to measure ICU length of stay. 

More recently, \cite{pawar2022leveraging} used a clinical language model in a multimodal setup for predicting 30-day all-cause mortality upon hospital admission in patients with COVID-19 and \cite{antikainen2023transformers} applied transformer models like BERT and XLNet on patient time series to predict 6-month mortality in cardiac patients. Advancing the integration, \cite{soenksen2022integrated} presented the Holistic AI in Medicine (HAIM) framework, outperforming similar single-source methodologies. Furthermore, \cite{bardak2021improving} integrated convolution over medical entities with multimodal learning for dual predictions. Despite these significant advancements, a gap remains in their practical applicability due to the moderate performance of the models. While benefiting from modern transformer-based models for text analysis and prediction tasks, the healthcare domain needs help with their real-world applicability. Our research emphasizes the power of transformer encoder models (particularly BERT and its derivatives) through ensemble modeling and text aggregation techniques, aiming to improve the models' performance and robustness.

\section{Methodology}

\begin{figure*}
	\centering
	
	\newlength{\subfigheight}
	\setlength{\subfigheight}{0.30\textheight}
	\addtolength{\subfigheight}{-3em}  
	
	\begin{subfigure}[b]{\textwidth}
		\centering
		\includegraphics[width=0.80\textwidth,height=\subfigheight]{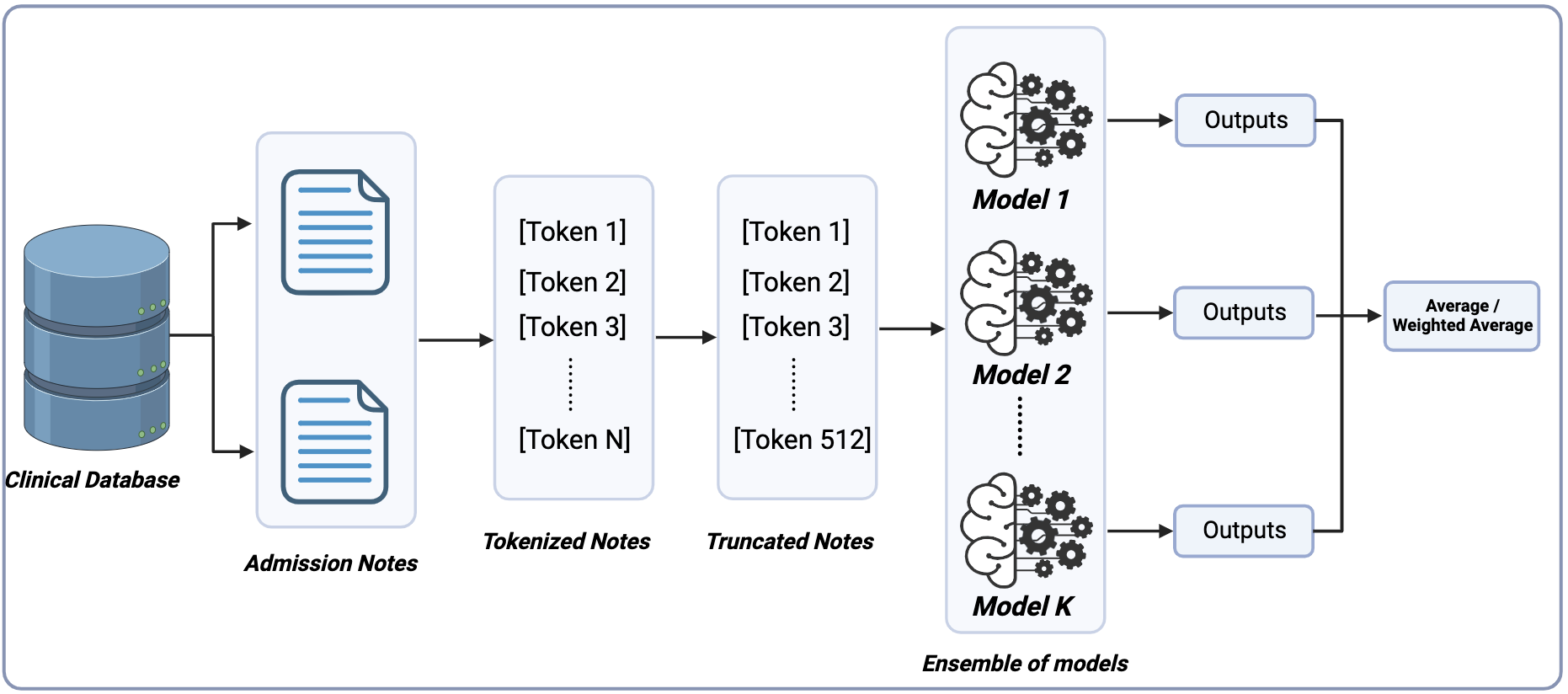}
		\caption{Ensemble: K instances of BERT-based models trained on clinical admission notes. The ensemble integrates these outputs.}
		\label{fig:sub1}
	\end{subfigure}
	\vspace{0.01em}
	
	\begin{subfigure}[b]{\textwidth}
		\centering
		\includegraphics[width=0.80\textwidth,height=\subfigheight]{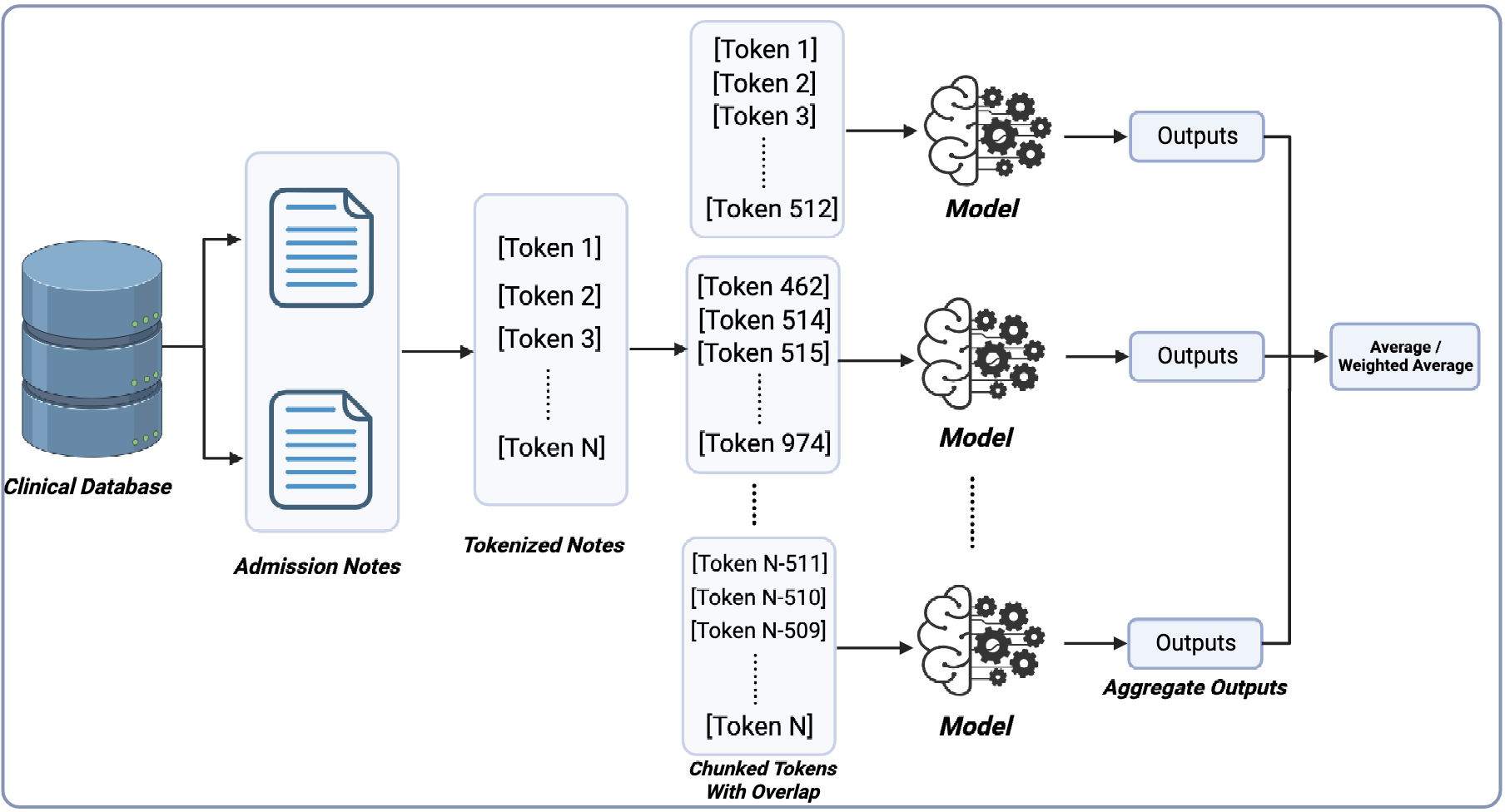}
		\caption{Text aggregation: Long clinical texts are divided into smaller segments, each one processed individually.}
		\label{fig:sub2}
	\end{subfigure}
	\vspace{0.01em}
	\begin{subfigure}[b]{\textwidth}
		\centering
		\includegraphics[width=0.80\textwidth,height=\subfigheight]{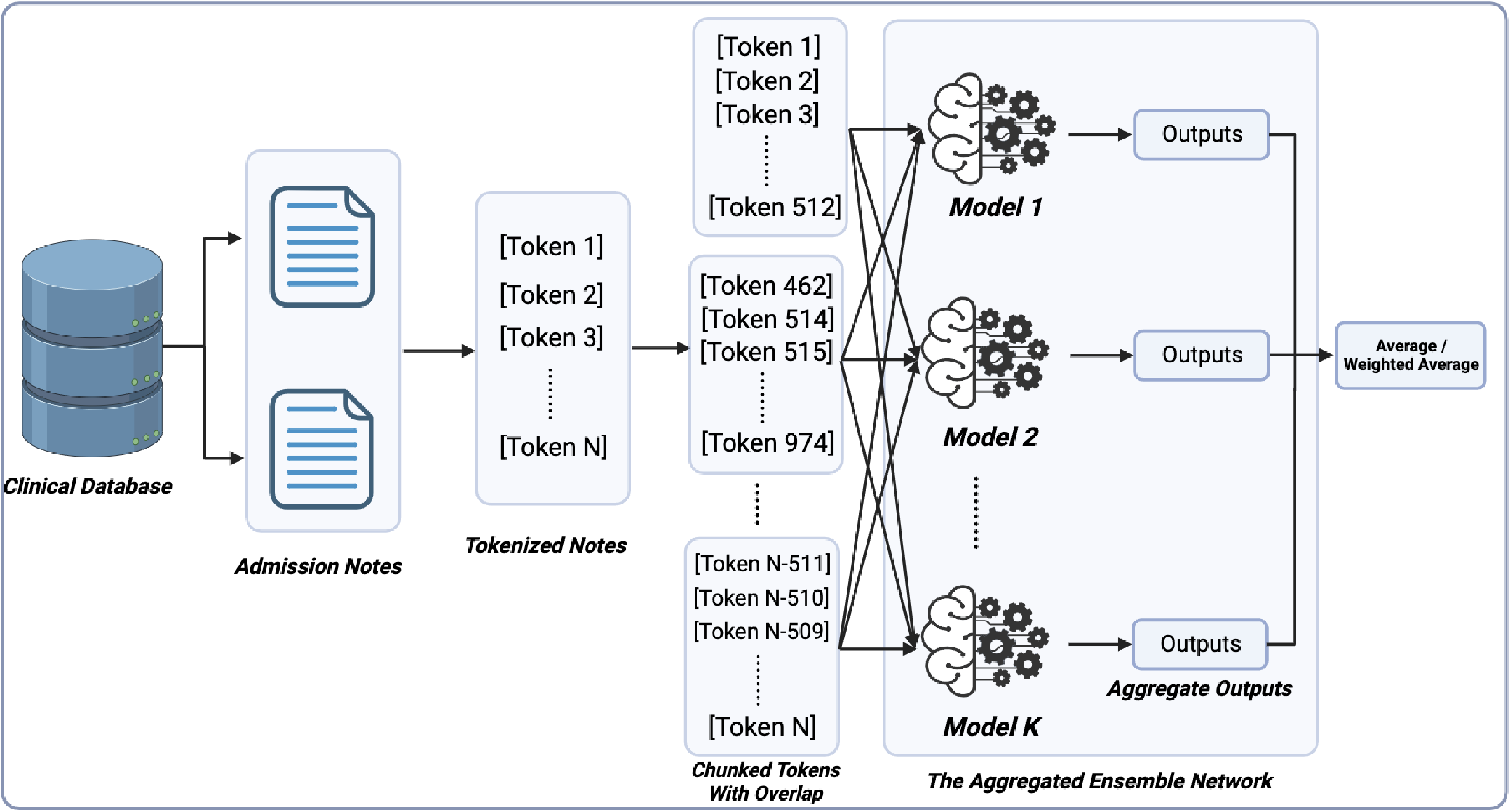}
		\caption{Ensemble and aggregation: Segmented long clinical notes are fed into an ensemble of BERT-based models. The collective predictions are then merged.}
		\label{fig:sub3}
	\end{subfigure}
	\caption{(a) Ensemble of multiple BERT-based models, (b) Aggregation process to handle long clinical texts, and (c) Unified ensemble and aggregation.}
	\label{fig:method_overview}
\end{figure*}

Using an ensemble-based approach coupled with strategic text aggregation, we aim to augment the strengths of multiple baseline models while restraining their limitations. Figure \ref{fig:method_overview} shows a cohesive representation of this design. The primary components encapsulated in the architecture are discussed below. 

\subsection{Baseline Models}

We used two pre-trained transformer-encoder models as the base models and then fine-tuned them using the dataset we discussed. As the pre-trained base models, we used CORe\cite{van2021clinical} and BioDischargeSummaryBERT\cite{alsentzer2019publicly}. These models used BERT and its spin-offs (pre-trained on related medical text) as base models.   

The CORe (Clinical Outcome Representations) \textit{CORe}  \cite{van2021clinical} used BioBERT \cite{lee2020biobert} (that used BERT \cite{devlin2018bert}) as the base model. According to the authors of \textit{CORe} \cite{van2021clinical}, this model undergoes a rigorous training on various data sources, from clinical notes and detailed disease descriptions to scholarly medical articles. Therefore, this model already embeds medical and clinical context, and we improve it more with the context of the MIMIC dataset.  

BioDischargeSummaryBERT \textit{DischargeBERT}            \cite{alsentzer2019publicly}: used ClinicalBERT \cite{clinicalBERT} as the base model. Therefore, this model embeds context from a different perspective. ClinicalBERT was pre-trained on a combination of clinical narratives and discharge summaries. However, DischargeBERT \cite{alsentzer2019publicly} elevates this by delving deeper into the nuances of the clinical domain.  

\subsection{Aggregation Method}

In the aggregation method, chunks of a clinical note is passed through a model to get the predictions, after which an average is taken to aggregate the outputs. Mathematically, 

\begin{itemize}
	\item \( C = \{c_1, c_2, ..., c_n\} \): Set of chunks obtained from tokenizing a clinical note.
	\item \( M(c_i) \): The prediction of the model \( M \) for chunk \( c_i \).
	\item \( P_M \): The aggregated prediction for the model \( M \).
\end{itemize}

The aggregation operation is fundamentally an averaging process:

\begin{equation}
	P_M = \frac{1}{n} \sum_{i=1}^{n} M(c_i)
\end{equation}

Here, \( P_M \) is the mean of the predictions across all chunks. By averaging over chunks, we ensure that each section of the note contributes equally to the final prediction regardless of its position. This approach is premised on assuming that all chunks hold potentially valuable information and should be treated equally.

\subsection{Ensemble Method}

With the goal that different models, while individually powerful, can have complementary strengths, we present a mathematical formulation of an ensemble of various models for improved generalization and performance. By integrating predictions across an ensemble of models, one can harness the collective intelligence, potentially reducing biases inherent to individual models and boosting prediction robustness.

Mathematically, we can denote the terms of formulation as follows:

\begin{itemize}
	\item \( E = \{M_1, M_2, ..., M_k\} \): Set of models in the ensemble.
	\item \( M_j(c_i) \): The prediction of the model \( M_j \) for chunk \( c_i \).
	\item \( P_E \): The ensemble prediction for all models in \( E \).
\end{itemize}

The ensemble prediction \( P_E \) is articulated as:

\begin{equation}
	P_E = \frac{1}{k} \sum_{j=1}^{k} \left( \frac{1}{n} \sum_{i=1}^{n} M_j(c_i) \right)
\end{equation}

In this equation, the inner summation computes the average prediction of model \( M_j \) across all chunks. On the other hand, the outer summation aggregates these averaged predictions across all models in the ensemble. Such a structure ensures the final prediction balances diverse model perspectives and a comprehensive aggregation of chunk-wise insights. The decision to average at the aggregation and ensemble level is rooted in statistical reasoning. As an ensemble method, averaging has proven effective in reducing variance while retaining the model bias. This ensures that the predictions are more stable and less sensitive to the nuances of individual data chunks or model-specific idiosyncrasies.

For the mortality prediction task, we fine-tune the models using the methodology for a binary classification task (0 for death and 1 for living), whereas, for the length of the prediction task, we fine-tune the models for a four-class classification task, where class 0 represents the length of stay $\leq$ 3 days, class 1 represents length of stay $>$ 3 days and $\leq$ 7 days, class 2 represents length of stay $>7$ days and $\leq$ 14 days. Lastly, class 3 represents length of stay $>$ 14 days.

\subsection{Algorithm}
Algorithm 1 shows the detailed steps described in the methodology. The goal is to efficiently utilize the power of the transformer-encoders for finding the context in long clinical texts.

\begin{algorithm}
	\label{alg: algo}
	\caption{Training algorithm of the research methodology}
	\begin{algorithmic}[1]
		\REQUIRE Clinical notes \( N \)
		\REQUIRE Ensemble of baseline models \( \mathcal{E} \)
		\REQUIRE Weight vector \( \mathbf{W} \) for aggregation
		\ENSURE Aggregated prediction
		
		\STATE Initialize an empty list: \( \mathcal{P} \)
		\STATE Define \( C = 510 \)
		
		\FOR{each note \( n_i \) in \( N \)}
		\STATE \( \mathcal{T} \leftarrow \text{tokenize}(n_i) \)
		\STATE \( \mathcal{C} \leftarrow \text{chunk} (\mathcal{T}, C) \)
		\FOR{each \( c_j \) in \( \mathcal{C} \)}
		\STATE Prepend [CLS] and append [SEP] to \( c_j \)
		\ENDFOR
		
		\COMMENT{Model Predictions}\\
		\FOR{each model \( m_k \) in \( \mathcal{E} \)}
		\FOR{each \( c_j \) in \( \mathcal{C} \)}
		\STATE \( o_{i,j,k} \leftarrow m_k(c_j) \)
		\ENDFOR
		\ENDFOR
		
		\COMMENT{Weighted Aggregation}\\
		\FOR{each \( c_j \) in \( \mathcal{C} \)}
		\STATE \( a_{i,j} \leftarrow \sum_{k} o_{i,j,k} \times W_k \) 
		\ENDFOR
		\STATE \( O_i \leftarrow \frac{1}{|\mathcal{C}|}\sum_{j} a_{i,j} \)
		\STATE Append \( O_i \) to \( \mathcal{P} \)
		\ENDFOR
		
		\RETURN \( \mathcal{P} \)
		
	\end{algorithmic}
\end{algorithm}

We aim to overcome the input length limitation of transformer encoder (for example, BERT) models. The main components we used to prove our point are fundamentally rooted in overcoming the intrinsic limitations of BERT-based models to process longer textual inputs, thereby maximizing the information retrieval from clinical notes.

\section{Experiments}

\subsection{Performance Metrics}

The primary metric adopted to measure the performance of our models is the Receiver Operating Characteristic Area Under the Curve (ROC-AUC). We used macro-averaged ROC-AUC to investigate the overall performance of our models comprehensively.

The ROC curve visualizes the trade-off between the true positive rate (TPR) and the false positive rate (FPR) across various threshold settings. Mathematically, it can be represented as:

\begin{equation}
	\text{TPR (Sensitivity)} = \frac{\text{TP}}{\text{TP} + \text{FN}}
\end{equation}

\begin{equation}
	\text{FPR (1 - Specificity)} = \frac{\text{FP}}{\text{FP} + \text{TN}}
\end{equation}

Where:
\begin{itemize}
	\item TP represents True Positives
	\item FN represents False Negatives
	\item FP represents False Positives
	\item TN represents True Negatives
\end{itemize}

The AUC provides a scalar value of the ROC's performance and represents the model's ability to distinguish between the classes. For multi-class problems, an individual ROC-AUC is computed for each class against the rest. The macro-average of the ROC-AUC is then calculated by averaging the individual ROC-AUCs. 

Let $AUC_i$ be the AUC for the $i^{th}$ class. For $n$ classes, the macro-averaged ROC-AUC is:

\begin{equation}
	AUC_{\text{macro}} = \frac{1}{n} \sum_{i=1}^{n} AUC_i
\end{equation}

This gives equal weight to each class, regardless of size, making it particularly useful when the dataset has class imbalances. In our experiments, given the multifaceted nature of clinical outcomes, the macro-averaged ROC-AUC provides a robust and unbiased performance assessment reflecting our models' discriminatory power.

\begin{table*}[ht]
	\centering
	\caption{Preliminary results for different experimented baseline models on mortality prediction and length of stay prediction tasks in macro-averaged \% AUROC. The CORe and DischargeBERT models outperform the baseline model performances, leading to their selection in the main experiments of our study.}
	\label{tab: preli results1}
	\begin{tabular}{lcc}
		\toprule
		\textbf{Model} & \textbf{Mortality Prediction} & \textbf{Length of Stay Prediction} \\
		\midrule
		BERT Base & 81.13 & 70.40 \\ 
		ClinicalBERT & 82.20 & 71.14 \\ 
		BioBERT Base & 82.55 & 71.59 \\ 
		\midrule
		CORe & 82.69 & 72.08 \\
		DischargeBERT & 82.32 & 72.24 \\ 
		\bottomrule
	\end{tabular}
\end{table*}

\begin{table*}[ht]
	\footnotesize
	\centering
	\caption{Preliminary results on mortality prediction and length of stay prediction tasks without text overlap in macro-averaged \% AUROC. The aggregated ensemble approach without text overlap outperforms other approaches, leading us to further experiment with text overlap.}
	\label{tab: preli results2}
	\begin{tabular}{llcc}
		\toprule
		\textbf{Category} & \textbf{Architecture} & \textbf{Mortality Prediction} & \textbf{Length of Stay Prediction} \\
		\midrule
		Baseline & CORe & 82.69 & 72.08 \\ 
		& DischargeBERT & 82.32 & 72.24 \\ 
		\midrule
		Ensemble & \shortstack{CORe + \\DischargeBERT} & 83.25 & 72.33 \\ 
		\midrule
		Aggregation & CORe & 83.64 & 72.53 \\
		& DischargeBERT & 84.27 & 72.60\\
		\midrule
		\shortstack{Ensemble + \\Aggregation} & \shortstack{CORe +\\DischargeBERT}  & \textbf{84.46} & \textbf{72.68} \\ 
		
		\bottomrule
	\end{tabular}
\end{table*}

\subsection{Implementation Details}

The following configurations and setups were used in the implementation of the experiments:

\begin{itemize}
	
	\item Hardware Configuration: Our computational platform consisted of an NVIDIA T4 Tensor Core GPU with 16GB of  memory. The cloud virtual machine contained 32 GB of RAM, ensuring smooth memory management during intensive computations.
	
	\item Training Epochs and Early Stopping: We trained models for up to 200 epochs. We also used an early stopping mechanism based on the ROC-AUC score as a monitor to avoid over-fitting. The training was terminated if the ROC-AUC did not improve by a delta of 0.0001 over three consecutive epochs.
	
	\item Batch Configuration: Given memory constraints and computational demands, we could use a batch size of 18 at max. Ensemble configurations adjusted this value based on the ensemble's size, determined by \( \frac{18}{\text{number of models in the ensemble}} \), ensuring uniform memory consumption distribution.
	
	\item Gradient Accumulation: To stabilize gradient updates and adapt to memory limitations, we used gradient accumulation - aggregating gradients over ten steps before executing an optimization iteration.
	
	\item Optimizer and Learning Rate Management: The AdamW optimizer was utilized, combining the advantages of Adam optimization and L2 weight decay. After comprehensive hyperparameter tuning, the learning rate was set at \(1 \times 10^{-5}\) with a weight decay of 0.01. Along with this, a linear warm-up learning rate scheduler was used. The initial 50 steps catered to learning rate adaptation, followed by a linear decay determined by \( \frac{\text{length of the train loader} \times \text{epochs}}{\text{gradient accumulation steps}} \).
	
	\item Model Preservation: In alignment with our focus on performance consistency, the iteration of the model yielding the highest ROC-AUC score during training was persistently saved, enabling seamless evaluation and future deployments.
	
\end{itemize}

\subsection{Training and Evaluation}

The foundation of any empirical study lies in its meticulous planning and rigorous experimentation, safeguarding against potential biases and ensuring reproducibility. Our methodology resonates with this ethos, prioritizing a robust training-validation-testing pipeline.

\begin{itemize}
	
	\item Data Splitting Strategy: We divided the total \textit{text} extracted from the notes into distinct train, validation, and test sets, as depicted in Table~\ref{tab:datasets}. This strategy was adopted to guarantee an unbiased evaluation, minimizing the risk of overfitting and ensuring a genuine representation of model performance.
	
	\item Training Protocols: We rigorously trained the models on the designated training set and optimized their weights based on performance metrics gleaned from the validation set. This iterative refinement assures convergence to a generalizable model proficient at handling unseen data variations.
	
	\item Evaluation and Reporting: We did the final evaluation on the untouched test set. Refraining from exposing the model to this dataset during training ensures an untainted appraisal of its capabilities, reflecting true external validity. We report the results of our experiments on this test set in the subsequent portions of the paper.
	
\end{itemize}

\subsection{Results}

\subsubsection{Preliminary Experiments}
\paragraph{\textbf{Experiment on different baselines}} Before diving into the findings and results of our primary experiments with the selected baseline models, it needs clarification as to why these models were specifically selected for experimentation and validation. Preliminary experiments on other baseline models like BERT-base, ClinicalBERT, and BioBERT-base revealed inferior performance on the clinical admission notes for the tasks of mortality prediction and length of stay prediction than CORe and DischargeBERT. Table \ref{tab: preli results1} shows these models' first preliminary experimental results. 

These results are consistent with the literature given that CORe underwent extensive pre-training using self-supervised clinical knowledge integration. Moreover, DischargeBERT was specifically pre-trained on the discharge notes from various clinical texts and knowledge bases, including the MIMIC-III clinical database. As a result, DischargeBERT and CORe outperformed all the other models in the clinical tasks of mortality prediction and length of stay prediction. We can verify this from Table \ref{tab: preli results1}.  

\paragraph{\textbf{Text Aggregation without Overlap}} We present the second preliminary results of this study in Table \ref{tab: preli results2}, where we experimented with the baseline models without text overlap in text aggregation. The baseline architectures, CORe and DischargeBERT, set the ground truth against which enhancements by subsequent methodologies are evaluated. 

The near-parity between the two baselines signifies that while both models have their strengths, neither possesses an overwhelming advantage in this application. 

Introducing the ensemble approach, which combines the power of both CORe and DischargeBERT, we observe a modest performance boost. The increment in performance underscores the complementary nature of the two models, with each compensating for the other's shortcomings. 

Delving into the aggregation method, where multiple chunks from the text are processed independently and their results averaged, we detect a notable improvement. This indicates the efficacy of the aggregation technique in harnessing more holistic insights from the data. 

We get the best results in the combined Ensemble + Aggregation approach. Merging the Ensemble's breadth with the aggregation's depth, the combined model realizes an apex AUROC of \(84.46\%\) for mortality prediction and \(72.68\%\) for the length of stay prediction. However, this motivated us to experiment further with aggregation methods using text overlapping since text overlapping can help preserve context between aggregation chunks and further improve the performance of the models.

\begin{table*}[ht]
	\footnotesize
	\centering
	\caption{Main experimental results on mortality prediction and length of stay prediction tasks with text overlapping in macro-averaged \% AUROC. The aggregated ensemble approach with text overlap outperforms all other approaches, including all preliminary results.}
	\label{tab: results_main}
	\begin{tabular}{llcc}
		\toprule
		\textbf{Category} & \textbf{Architecture} & \textbf{\shortstack{Mortality \\Prediction}} & \textbf{\shortstack{Length of Stay \\Prediction}} \\
		\midrule
		Ensemble & \shortstack{CORe +  \\ DischargeBERT} & 83.25 & 72.33 \\ 
		\midrule
		\shortstack{Aggregation \\(with overlap)} & CORe & 83.21 & 72.33 \\
		& DischargeBERT & 84.50 & 72.62\\
		\midrule
		\shortstack{Ensemble + \\Aggregation (with overlap) }& \shortstack{CORe +  \\ DischargeBERT} & \textbf{84.52} & \textbf{72.78} \\ 

		\bottomrule
	\end{tabular}
\end{table*}

\begin{figure*}[ht]
	\centering
	\includegraphics[width=\linewidth]{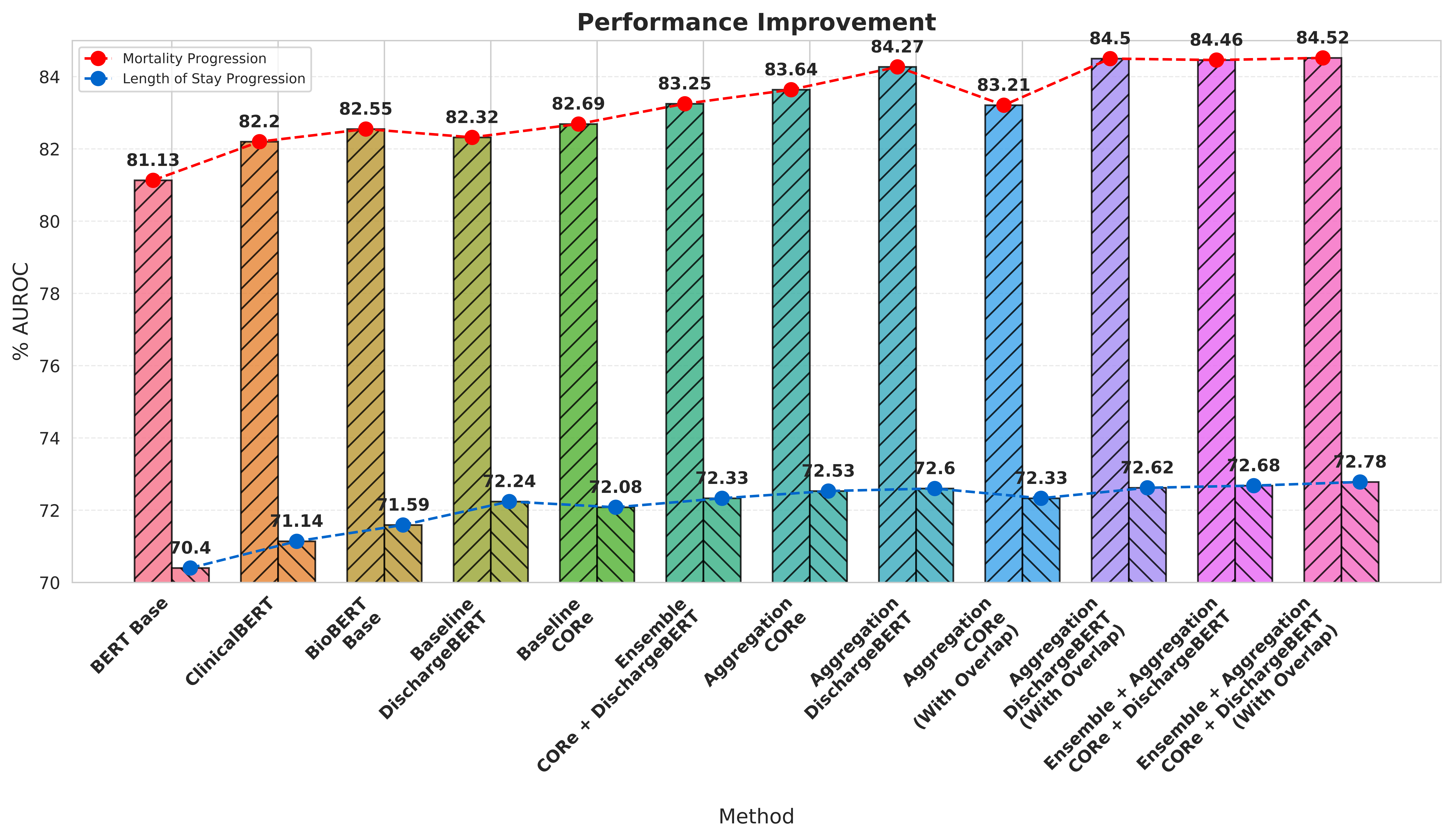}
	\caption{Comparative analysis and gradual performance improvement from preliminary approaches to selected baselines, ensemble models, aggregation models without overlap, aggregation models with overlap, Ensemble + Aggregation without overlap, and finally, Ensemble + Aggregation with overlap, which gives the best results in macro averaged \% AUROC for the clinical tasks Mortality Prediction and Length of Stay Prediction.}
	\label{fig:progress_all}
\end{figure*}

\paragraph{\textbf{Summarization on Clinical Notes}} We further experiment with extractive 
and abstractive summarization of the clinical admission notes. Extractive and abstractive summarization can extract the most essential and relevant information from long texts to condense into a smaller, more informed piece of text \cite{zhong2020extractive}.

However, the domain of clinical NLP is not enriched with pre-trained summarization models. Due to the unstructured, irregular, and sparse nature of clinical texts, powerful summarization models like BART \cite{lewis2019bart} have proven to perform poorly on the summarization of these texts. Moreover, abstractive summarization models like \cite{zhang2020pegasus} and extractive summarization models like Longformer-sum have also proven ineffective in summarizing and condensing the most relevant medical information under 512 tokens.

More crucial reasons behind the poor summarization of these models are that the clinical texts are enriched with numerical data and negated terms, which the models need help comprehending or interpreting as required. Hence, our preliminary experiments using these methods unveiled a significant requirement of summarization model pre-training on clinical data. For future work, we suggest summarization model pre-training and generative large language models like GPT \cite{brown2020language} to condense long clinical texts into smaller, more informative ones.

\subsection{Main Experiments and Results}

The main findings of this study are shown in table \ref{tab: results_main}, where we perform text aggregation with an overlap of 50 tokens to increase coherence and preserve context between text chunks. 

A comparative analysis between the preliminary results and the main results shows that text aggregation with overlapping improves the performance of the models for both the clinical tasks: mortality prediction and length of stay prediction. This is observed in the case of both aggregation and Ensemble + Aggregation approaches. 

This dual approach capitalizes on the models' individual strengths and seamlessly integrates them, leading to a synergistic elevation in performance. Figure \ref{fig:progress_all} further shows a comparative visualization of the performance of the models, where the graph extends towards the Ensemble + Aggregation with overlap method better than the other methods, showing the best performance.

\section{Discussion and Findings}
Based on the experimental results of the previous section, here we present some keen observations and limitations of our method:

\paragraph{\textbf{Aggregation is better than ensembles}} The aggregation method is more effective than the ensemble method in improving the performance of both CORe and DischargeBERT. This suggests that the long input texts contain valuable information lost or diluted when only a single chunk is processed. By averaging the results from multiple chunks, the aggregation method can capture more comprehensive and accurate insights from the data.

\paragraph{\textbf{Complementary nature of baseline models}} The combined Ensemble + Aggregation approach achieves the best results for mortality and length of stay prediction tasks. This demonstrates that the two models, CORe and DischargeBERT, have complementary strengths and weaknesses that can be leveraged by combining them. The ensemble method can exploit the diversity of the models, while the aggregation method can enhance the robustness of the models.

\paragraph{\textbf{Task-based performance variation}} The difference between the performance of CORe and DischargeBERT is larger for mortality prediction than for length of stay prediction. This gap implies that DischargeBERT has a more potent ability to handle the mortality prediction task, which may require more semantic understanding and reasoning skills. CORe, on the other hand, is more suitable for the length of stay prediction task, which may depend more on the syntactic and structural features of the text.

\paragraph{\textbf{Mortality prediction vs. length of stay prediction}} The improvement in performance by using the Ensemble + Aggregation approach is more significant for mortality prediction than for length of stay prediction. This indicates that the mortality prediction task is more challenging and complex than the length of stay prediction task and thus benefits more from the fusion of different models and data sources. The length of stay prediction task may have a lower ceiling for performance improvement, as other factors beyond the text input may influence it.

\paragraph{\textbf{Limitations and future work}} Despite the promising findings of our research and the robust performance of the models in predicting clinical outcomes, the following concerns remain:

\begin{itemize}
	\item The computational overhead of training the models using model ensemble and text aggregation requires clarity. Deploying these models in clinical healthcare systems would require less inference time for the models, irrespective of the training duration. Although the inference times by the trained models are suitable for application in current healthcare systems, there is room for ample improvement. This can be achieved with model optimization techniques such as knowledge distillation \cite{hinton2015distilling, guo2023class, ji2021show}, model quantization\cite{jacob2018quantization, kim2021bert} or pruning methods\cite{chen2020lottery, mccarley2019structured}. For future works, we suggest focusing on these techniques to reduce the inference time of the models.
	
	\item Our aggregation method effectively segments long clinical texts into meaningful chunks with a certain overlap. Even though rigorous experimentation and hyperparameter optimization techniques accomplished the selection of the number of tokens for the overlap, there might be room for even better context preservation using techniques like attention mechanisms or hierarchical transformers. However, these are still active research areas and require exploration and further focus.
\end{itemize}

\section{Conclusion}

In this research, we aimed to process medical texts using transformer encoders without truncating. Transformer encoders put a limit on how many tokens it can process at the same time to find the context. If we want to process medical text partially due to these limitations, we risk losing essential knowledge that may otherwise provide keen insights into critical medical processes. 

We demonstrate our scheme using various BERT-based encoders and clinical admission notes from the MIMIC-III clinical database. We found that integrating model ensembling with text aggregation robustly enhances transformer-encoder-based models' performance in predicting mortality and length of stay. Our approach addresses the limitation of handling long clinical texts and outperforms baseline benchmarks and individual strategies. This method promises to significantly benefit clinical healthcare systems by enabling comprehensive processing of diversified data sources, leading to more informed decisions and improved patient outcomes. As medical informatics progresses in the era of data-rich healthcare, our findings highlight the immense potential of aggregated ensembles of large language models in optimally harnessing clinical knowledge.

\section*{Ethics and Consent}

\subsection*{About the Dataset}
The data that support the findings of this study are available from the MIMIC-III (Medical Information Mart for Intensive Care III) v1.4 database \cite{johnson2016mimic}. The dataset has a DOI of 10.13026/C2XW26 and can be accessed at https: //mimic.physionet.org/. The dataset is licensed under the ODC Open Database License v1.0. Researchers seeking to use the database must formally request access and complete a required training course. The part of data used in this study is de-identified, and requires no permission from the actual subjects.

\subsection*{Ethical Approval}
The research involving the data of the MIMIC-III clinical database is approved by the Partners Human Research Committee, with the IRB number 2001-P-001699/14.

\section*{Acknowledgement}
This work was supported by the North South University Office of Research [grant number CTRG 2020-2021/CTRG-20-SEPS-02].


\end{document}